\documentclass{article}

\pdfoutput=1

\usepackage[preprint]{neurips_2023}

\usepackage[utf8]{inputenc} 
\usepackage[T1]{fontenc}    
\usepackage{hyperref}       
\usepackage{url}            
\usepackage{booktabs}       
\usepackage{amsfonts}       
\usepackage{nicefrac}       
\usepackage{microtype}      
\usepackage{xcolor}         
\usepackage{graphicx}

\usepackage{natbib}
\setcitestyle{comma}
\setcitestyle{numbers}
\setcitestyle{square}

\title{Quantifying Overfitting: Evaluating Neural Network Performance through Analysis of Null Space}

\author{%
  Hossein~Rezaei \\
  School of Computer Engineering\\
  Iran University of Science and Technology (IUST)\\
  \texttt{hossein\_rezaei@comp.iust.ac.ir} \\
  \And
  Mohammad Sabokrou$^{1,2}$ \\
   $^1$Okinawa Institute of Science and Technology\\
   $^2$Institute For Research In Fundamental Sciences \\
  \texttt{mohammad.sabokrou@oist.jp} \\
}

\begin{document}

\maketitle

\begin{abstract}
  Machine learning models that are overfitted/overtrained are more vulnerable to knowledge leakage, which poses a risk to privacy. Suppose we download or receive a model from a third-party collaborator without knowing its training accuracy. How can we determine if it has been overfitted or overtrained on its training data? It's possible that the model was intentionally over-trained to make it vulnerable during testing. While an overfitted or overtrained model may perform well on testing data and even some generalization tests, we can't be sure it's not over-fitted. Conducting a comprehensive generalization test is also expensive. The goal of this paper is to address these issues and ensure the privacy and generalization of our method using only testing data. To achieve this, we analyze the null space in the last layer of neural networks, which enables us to quantify overfitting without access to training data or knowledge of the accuracy of those data. We evaluated our approach on various architectures and datasets and observed a distinct pattern in the angle of null space when models are overfitted. Furthermore, we show that models with poor generalization exhibit specific characteristics in this space. Our work represents the first attempt to quantify overfitting without access to training data or knowing any knowledge about the training samples. \footnote{
The source code will be available after the review.}
\end{abstract}

\section{Introduction}
Deep learning models have been very successful in many applications such as computer vision, natural language processing, and speech recognition. These models are trained on large amounts of data and have demonstrated outstanding performance in tasks such as image classification, object detection, and language translation \citep{sarker2021machine, zhao2019object}. However, despite their effectiveness, ensuring the privacy and trustworthiness of deep learning models remains a significant challenge \citep{li2022interpretable, zhao2019privacy, mireshghallah2020privacy}.

\begin{figure}[t!]
  \centering
  \includegraphics[width=\textwidth]{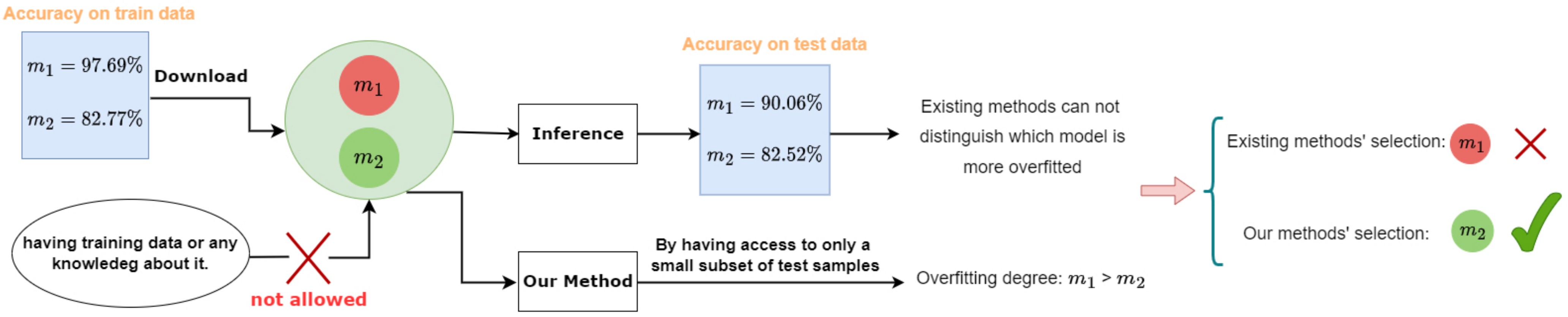}
  \caption{We possess two downloaded models with no information regarding their training data or training accuracy. Additionally, we only have access to a limited subset of test data. Despite model 1 exhibiting higher test accuracy, it appears to suffer from overfitting, and existing current methods are unable to effectively discern this issue.}
  \label{fig:Intro}
\end{figure}

In today's data-driven world, accessing pre-trained models has become increasingly common, whether obtained from the internet or delivered by third-party companies. However, it is crucial to ensure that these models uphold privacy standards and do not possess knowledge leakages. A key factor in determining the vulnerability of a model to membership inference attacks is the presence of overfitting. Generally, the more overfitted a model is, the more susceptible it becomes to such attacks. However, assessing this characteristic becomes challenging when we lack information about the model's training accuracy or training data. In this paper, we aim to address this critical question and explore potential solutions for evaluating model vulnerability in situations where these crucial details are unavailable (see Fig.~ \ref{fig:Intro}).

Generally, one of the key concerns in deep learning is that the models often memorize the training data \citep{caruana2000overfitting, ying2019overview}. This means that the models may overfit the training data, resulting in poor generalization to new data. Additionally, the models may memorize sensitive information from the training data, which can pose a risk to privacy \citep{yeom2018privacy}. For instance, if a model has a privacy leakage, attackers may be able to extract sensitive information from the model during inference \citep{jagielski2022measuring, carlini2019secret}. If an attacker gains access to such information, it can have severe consequences for individuals or organizations.

A primary factor that makes a deep learning model vulnerable to privacy breaches is overfitting \citep{jagielski2022measuring}. Researchers have proposed various methods to address this issue and enhance the privacy and trustworthiness of deep learning models. For example, \citep{ghazi2021deep, malek2021antipodes} leveraged differentially private training, \citep{thakkar2020understanding, huang2022detecting} exploited gradient clipping, and \citep{neel2021descent, sekhari2021remember} utilized machine unlearning to improve privacy and prevent knowledge leakage.

One of the simplest ways to detect overfitting is by comparing the accuracy of the model on the training and testing datasets. If the model achieves high accuracy on the training data but low accuracy on the testing data (low bias and high variance), it may be overfitting \citep{ying2019overview, geman1992neural}. However, obtaining the accuracy of the training data requires access to the training dataset, which may not always be feasible or ethical. Some papers, like \citep{shokri2017membership, carlini2019secret, jagielski2022measuring}, attempt to measure forgetting and memorization by conducting attacks while relying on training data to accomplish this task.

Another approach to detecting overfitting is by performing a generalization/robustness evaluation \citep{novak2018sensitivity, schiff2021predicting, jiang2021methods}. This test evaluates how well the model can generalize to new data by measuring its performance on a separate dataset that it has not seen before. If the model performs well on the generalization test, it is less likely to suffer from overfitting. However, this approach has some drawbacks. Firstly, it can be costly to implement, as it requires collecting several separate datasets for the generalization test and it takes too much time to perform multiple inferences from the model. Secondly, since generalization tests are widely known, an attacker could  overtrain/overfit the model on those tests, making the model robust to those specific tests and potentially opening the door for privacy breaches.

In addition to the aforementioned methods, another simple approach to investigate the issue of overfitting is to examine the uncertainty of the model by analyzing the soft-max output or logits \citep{hong2018overfitting}. The idea is that probably there is a direct relationship between the uncertainty of the model and the overfitting. However, in Section \ref{Results}, we demonstrate that this argument does not always hold true.

To address these challenges,  we propose a novel method to detect overfitting and ensure the privacy and generalization of deep learning models using only a small amount of test data. The proposed method involves analyzing the null space in the last layer of neural networks, which enables us to quantify overfitting without access to the training data or knowledge of its accuracy. The null space is the set of all vectors that the neural 
 network maps to zero. We interestingly find that by analyzing the angle between the null space of weights and the representation,  can give us supervision to detect overfitting and determine the generalization performance of the model. The proposed method has been evaluated on various architectures and datasets, and the results show that there is a distinct pattern in the angle of the null space when models are overfitted. Furthermore, we illustrate that models exhibiting poor generalization display specific characteristics within this space. \textit{The proposed method  represents one of the first attempts to quantify overfitting without access to training data or any knowledge about the training samples}. The method is easy to implement and can be applied to various architectures and datasets, making it a promising tool to enhance the privacy of deep learning models.

\section{Related Work}

In this section, we provide a review delving into the literature attempting to measure the overfitting and generalization capability of machine learning models. Furthermore, we explore several works that leverage null space across various applications using neural networks.

\subsection{Overfitting \& Generalization}

Overfitting arises when a model becomes too complex and memorizes the training data instead of learning the representative patterns, resulting in failure to generalize well to unseen datasets. To address this issue, Werpachowski et al.~\citep{werpachowski2019detecting} introduce a non-intrusive statistical test using adversarial examples to detect test set overfitting in machine learning models. Yet, one notable challenge they highlighted is accurately measuring test set overfitting due to shifts in data distribution.
Moreover, Jagielski et al.~\citep{jagielski2022measuring} explored memorization, forgetting, and their impact on overfitting via a privacy attack method. They train two models with extra data to measure forgetting, using the success rate to identify retained sensitive information and discarded irrelevant or noisy information.
Carlini et al. \citep{carlini2019secret} consider a testing approach that evaluates the level of risk associated with generative sequence models inadvertently memorizing infrequent or distinct training data sequences. To assess the level of overfitting in convolutional neural networks (CNNs), PHOM~\citep{watanabe2022overfitting} employs trained network weights to create clique complexes on CNN layers. By examining co-adaptations among neurons via one-dimensional persistent homology (PH), it detects overfitting without relying on training data. PHOM differs from our work in terms of efficiency and complexity.

Generalization and Out-Of-Distribution (OOD) generalization refer to our model's ability to adapt appropriately to new, previously unseen data drawn from either the same distribution or a different distribution as the training data, respectively. To address this issue, Neyshabur et al.~\citep{neyshabur2017exploring} connect sharpness to PAC-Bayes theory and show that expected sharpness, a measure of network output change with input change, along with weight norms, can capture neural network generalization behavior effectively. Some other works attempt to evaluate the generalization of deep networks by defining bounds. For instance, Liang et al.~\citep{liang2019fisher} introduce the Fisher-Rao norm, an invariant measure based on information geometry. It quantifies the local inner product on positive probability density functions (PDFs) and relates the loss function to the negative logarithm of conditional probability, with Fisher information as the gradient.
Kuang et al. \citep{kuang2020stable}, and Shen et al. \citep{shen2020stable} use the average accuracy to measure OOD generalization. While Duchi et al. \citep{duchi2018learning}, and Esfahani et al. \citep{mohajerin2018data} measure the OOD generalization using worst-case accuracy.

Unlike current approaches, we propose to measure overfitting and generalization without access to the training data or training accuracy, utilizing only a small subset of the test set to determine the degree of overfitting and generalization capability.

\subsection{Null Space}
The concept of the null space is important across diverse domains of mathematics, such as linear algebra, differential equations, and control theory. In the context of neural networks, the null space of a weight matrix can be used for various applications. Most research work is focused on analyzing null space for out-of-distribution detection.

In novelty detection, Bodesheim et al.~\citep{bodesheim2013kernel} use null space for detecting samples from unknown classes in object recognition by mapping training samples to a single point, enabling joint treatment of multiple classes and novelty detection in one model. IKNDA~\citep{liu2017incremental} addressed the demanding computational burden caused by kernel matrix eigendecomposition in this method and performed novelty detection by extracting new information from newly-added samples, integrating it with the existing model, and updating the null space basis to add a single point to the subspace. For outlier detection, Null Space Analysis (NuSA)~\citep{cook2020outlier} is proposed to detect outliers in neural networks using weight matrix null spaces in each layer. It provides competency awareness in ANNs and tackles adversarial data points by controlling null space projection. 
Likewise, Wang et al.~\citep{wang2022vim} utilize null space to measure out-of-distribution (OOD) degree. by decomposing feature vectors, generating confident outlier images, and subsequently calculating angle-based OOD score.
Additionally, Idnani et al.~\citep{idnani2023dont} explore the null space's impact on OOD generalization, introducing null space occupancy as a failure mode in neural networks. They optimize network weights using orthogonal gradient descent to reduce null space occupancy, which enhances generalization.

Drawing inspiration from the application of null space in out-of-distribution detection, we employ null space properties to assess both the degree of overfitting and the generalization capacity of machine learning models.

\section{Proposed Method}
\label{headings}
The aim of this study is to explore how to detect overfitting in machine learning models without prior knowledge of the training samples or accuracy. We discovered a close correlation between the weights associated with each class and the representation. If the weight for each class is orthogonal to the representation, it means that the input does not belong to that class. Conversely, if the weight is in the same direction as the representation, the angle between them is close to zero, indicating that the input belongs to that class. Although the main concepts of different classes are different, there are some common/shared characteristics between them. Therefore, the angle between the representation and the weight of targeted class should be close to zero, but there should be some gap/angles that reflect the relationship with other classes.
We found that when a model is over-fitted or over-trained, it loses of relationship with other classes (the angle between the target class weight and the representation is very close to zero) and less generalization, which can result in overfitting.
To apply these findings, we propose monitoring the angle between the weights and the representation during model training. 
Our approach provides a simple and effective method for detecting overfitting in machine learning models. By using our proposed method, we can detect models that generalize well to new data and avoid overfitting, even without prior knowledge of the training data or accuracy.

 Formally speaking, We investigate the concepts of overfitting  from a null space perspective. As mentioned, the  goal is to determine whether models are overfitted or not and analyze their generalization capability. To accomplish this, suppose $\mathcal{M}$ is a set of models \begin{math}\mathcal{M} = \{m_1, m_2,..,m_{k}\}\end{math}, and access the  test data (or validation samples) \begin{math}\mathcal{X} = \{x_1, x_2,..,x_n\}\end{math}.  \begin{math}\mathcal{X}\end{math} feeds into the network aiming to represnt it i.e., \begin{math}\mathcal{R} = \{r_1, r_2,..,r_n\}\end{math}  ($r_i$ corresponds  the $x_i$ samples). Then we leverage the angle between \begin{math}r_i\end{math} and the null space of weights that are not associated with ground truth, as well as the angle between \begin{math}r_i\end{math} and the weights associated with ground truth, to establish two scores for measuring overfitting and generalization. These Scores are defined as follows:

 \[\mathcal{O} = \alpha + \beta,\]
 \[\mathcal{G} = \frac{\alpha}{max(\alpha)} + \frac{|\beta|}{max(|\beta|)},\]

Where, \begin{math}\mathcal{O} \end{math} denotes the degree of overfitting, while \begin{math}\mathcal{G} \end{math} represents the amount of generalization capability. \begin{math} 
\alpha\end{math} denotes the average of the angles between the \begin{math}\mathcal{R}\end{math} and weight vector of target classes, while \begin{math}\beta\end{math} defines the average angles between \begin{math}\mathcal{R}\end{math} and the null space of weight vectors (column space) of false classes.

\begin{figure}[t!]
  \centering
  \includegraphics[width=\textwidth]{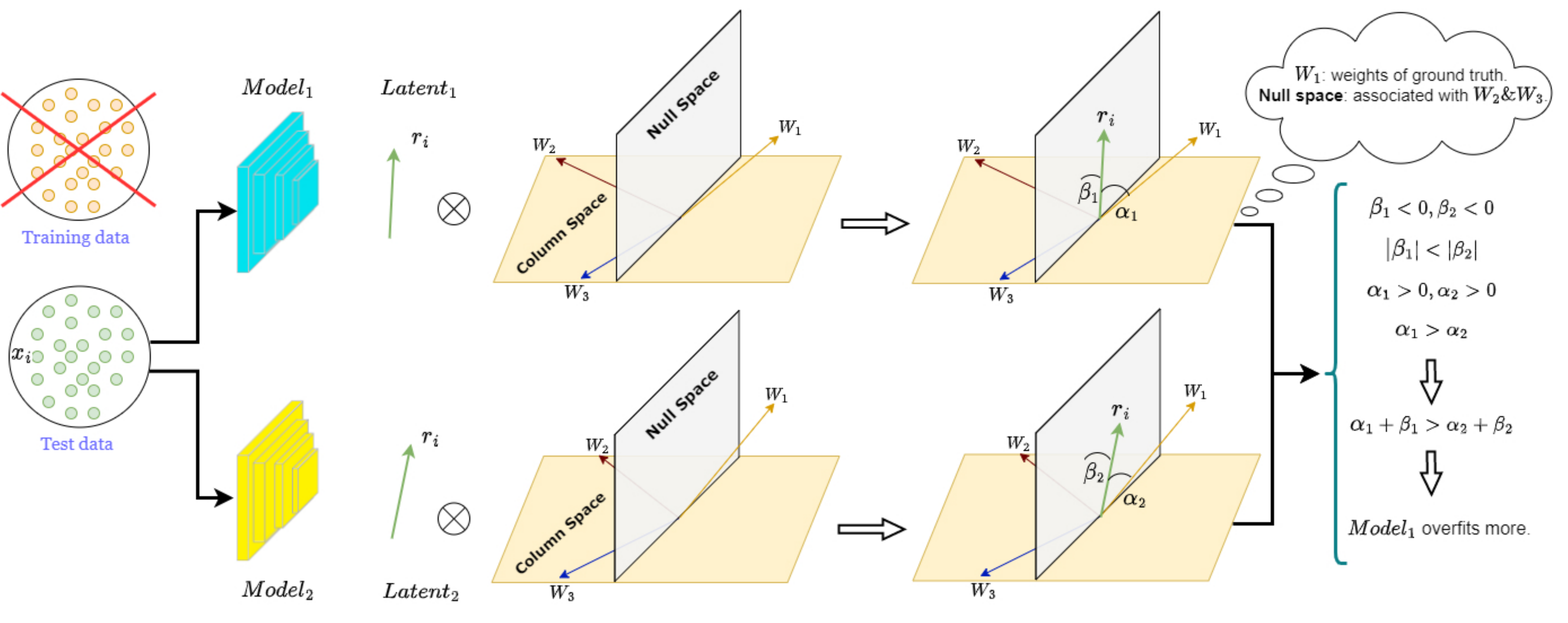}
  \caption{Our method is based on a simple framework. We randomly select a subset of the test data to compare the degree of overfitting of the two models. The weights vector of the target class is denoted by \begin{math}W_1\end{math}, while the Null Space plane corresponds to the Null Space associated with the weights vectors of the false classes (\begin{math}W_2 \& W_3\end{math}). To compute the degree of overfitting, we first pass the test samples through the encoder to obtain their representation vectors. Next, we measure the angle between the representation vector and the null space plane (\begin{math}\beta\end{math}) and the angle between the representation vector and the true class weight vector (\begin{math}\alpha\end{math}) ( we perform this process for all samples and ultimately calculate the average). Finally, we compute the sum of \begin{math}\alpha\end{math} and \begin{math}\beta\end{math}, which serves as a quantitative measure of the degree of overfitting.}
  \label{fig:Angles}
\end{figure}

\subsection{Null Space \& Column Space}

In linear algebra, the null space and column space are two fundamental subspaces associated with a matrix. The null space is sometimes called the kernel of the matrix, while the column space is sometimes called the range of the matrix. The null space of an m × n matrix A is a subspace of \begin{math}\mathcal{R}^n\end{math}, written as \begin{math}Nul(A)\end{math}, and defined by:

\[Nul(A) = \{ x\in\mathcal{R}^n \, |\, Ax = 0\}\]

Where A refers to a linear mapping. Geometrically, the null space represents all the directions in which the matrix A "collapses" to zero. The column space of an m × n matrix A, written as \begin{math}Col(A)\end{math}, is a subspace of \begin{math}\mathcal{R}^m\end{math}, and is the set of all linear combinations of the columns of A. In other words, The column space of a matrix A is the span of its columns. Geometrically, the column space represents the "shadow" of the matrix A, as cast onto a lower-dimensional subspace. If \begin{math} A = [a_1,...,a_n],  \end{math} then

$$Col(A) = Span\{a_1,...,a_n \} $$

Now, it is true that the null space and column space of a matrix are orthogonal complements of each other. This means that any vector in the null space is orthogonal to any vector in the column space, and vice versa. In fact, the left null space is orthogonal to the column space of A. To see why this is true, consider the Appendix \hyperref[AppendixA]{A}.

\subsection{Over-fitting \& Generalization Measurement }
To analyze overfitting, we split the weight vectors into two groups: The first group for the weights vector of the true class (target class), and the second group for the weights vectors of the false classes. We then analyze the behavior of the representation vector toward these two groups.

\paragraph{Null Space Angle.}
In deep learning models, we use the inner product. Specifically, we use the following formula:

\[y = w^T\cdot x\]

where \begin{math}<\cdot>\end{math} represents the inner product between the representation vector and the transpose of weight vectors, and y represents the logits. Since the vectors in the second group represent false classes, their logits should have low values and should not significantly influence the output decision. Therefore, the angle between the weight vectors of false classes (group 2) and \begin{math}x\end{math} should be close to 90 degrees. Furthermore, as discussed in \citep{wang2022vim}, the dimensions of the representation vector are typically larger than the dimensions of the logits. This can result in some information loss when the representation vector is fed into the MLP layers. By leveraging the null space, and the behavior of the representation vector toward this space we can potentially recover some of this lost information, which may be useful for analyzing overfitting and generalization.

The space spanned by the vectors in the second group is known as the column space. As previously mentioned, the null space is orthogonal to this space. For the aforementioned reasons, to analyze the relationship between the representation vector and the vectors in the second group, we utilize the concept of null space. In other words, we measured the angle between the null space and the representation vector. In this way, We found that this angle provides us with useful information for analyzing overfitting. In  Fig. \ref{fig:Angles}, \begin{math}\beta\end{math} represents this angle.

\paragraph{True Angle}

As previously discussed in the Null Space Angle section, deep learning models use the inner product to predict the output. Since the output of the network depends on the argmax of Logits/SoftMax, the corresponding logit for the inner product of the representation vector and the vector from group 1 (the weight vector of the target class) should have the maximum value among the other logits. To ensure this, the angle between the representation vector and the vector from group 1 should be close to zero.

We analyzed this angle and found that it provides us with some information about overfitting and generalization. Therefore, we measured it to determine the degree of overfitting and generalization capability. In Fig. \ref{fig:Angles}, \begin{math}\alpha\end{math} represents this angle.

\paragraph{Overfitting}

We have observed that when the network exhibits good forgetting (the network was not overfitted), it tends to optimize two things simultaneously. Firstly, it minimizes the angle between the representation vector and the correct class weight vector (i.e., the vector associated with group 1) to ensure that the representation vector to what extent is aligned with the correct class. Secondly, it maximizes the angle between the representation vector and the null space.

In other words, the network is trained to adjust its learnable weights and parameters in a way that moves the representation vector away from the null space while simultaneously maximizing its projection onto the vector of group 1. This process leads to a decrease in the value of \begin{math}\alpha\end{math} and an increase in the absolute value of \begin{math}\beta\end{math} (or a decrease in \begin{math}\beta\end{math} itself), resulting in an overall decrease in the sum of \begin{math}\alpha\end{math} and \begin{math}\beta\end{math}. \begin{math}\alpha\end{math} represents to what extent the network correctly predicts the label, while \begin{math}\beta\end{math} indicates how likely the network considers the representation vector to be similar to other classes.

Therefore, the sum of \begin{math}\alpha\end{math} and \begin{math}\beta\end{math} serves as an indicator of the degree of overfitting. The lower this value is, the less overfitting the model is.

\textbf{Important note:} In fact, the angle between the representation vector and the weights vector of the target class (denoted as \begin{math}\alpha\end{math} in Fig. \ref{fig:Angles}) indicates the relationship between the input image and the target class. Meanwhile, the angle between the representation vector and the null space (denoted as \begin{math}\beta\end{math} in Fig. \ref{fig:Angles}) reflects the average behavior of the input image towards false classes. For instance, let's consider an example using CIFAR10, where our input sample is a cat. In this case, the representation vector should be close to the weights vector of the cat target class. Furthermore, since there is another class called "dog" in CIFAR10, the representation vector of the cat should be slightly closer to the weights vector of the "dog" class (since the cat and dog bear some resemblance to each other; slightly less than 90 degrees).

On the other hand, classes like "ship" and "truck" have no similarity to the "cat" class. Hence, the angle between the representation vector and the weights vectors of these classes should be slightly greater than 90 degrees. It is worth noting that the angle between the representation vector and the null space (denoted as \begin{math}\beta\end{math} in Fig. \ref{fig:Angles}) represents the average behavior of the input image towards false classes. Therefore, on average, the angle between the representation vector and this space should be more than 90 degrees. Due to this reason, as this angle is greater than 90 degrees, it falls on the right side of the space, and we consider this angle as negative.

\paragraph{Generalization}
We have discovered that when the network possesses good generalization capabilities, it attempts to reduce the projection of the representation vector onto both the null space and the weight vector of the target class. This behavior resembles the concept of overfitting, wherein the network strives to move the representation vector away from the null space. However, in terms of moving the representation vector away from the weight vector of the target class (true class), it opposes overfitting. It is important to note, as demonstrated by the results of various corruptions (as shown in Fig. \ref{fig:Generalization}), that a model with less overfitting does not necessarily exhibit superior generalization ability. Hence, we have come to understand that if a model aims to possess both high generalization ability and reduced overfitting While increasing the angle of the representation vector with the null space, it should establish a balance between a small or large angle of the representation vector and the weights vector of the target class. Essentially, the projection should neither be excessively high nor too low.

\section{Experiments}
\label{others}

We evaluate the performance of our method on several widely-used convolutional neural network architectures, including ResNet18, ResNet34, ResNet50, DenseNet121, VGG19, and MobileNetV2, using three different datasets: CIFAR10, SVHN, and CIFAR100. To conserve space, we present the results for the ResNet18 architecture on CIFAR10 in the main text, with additional experiments provided in Appendix \hyperref[AppendixBC]{B}.

\subsection{Setup}

\begin{figure}
  \centering
  \includegraphics[width=\textwidth]{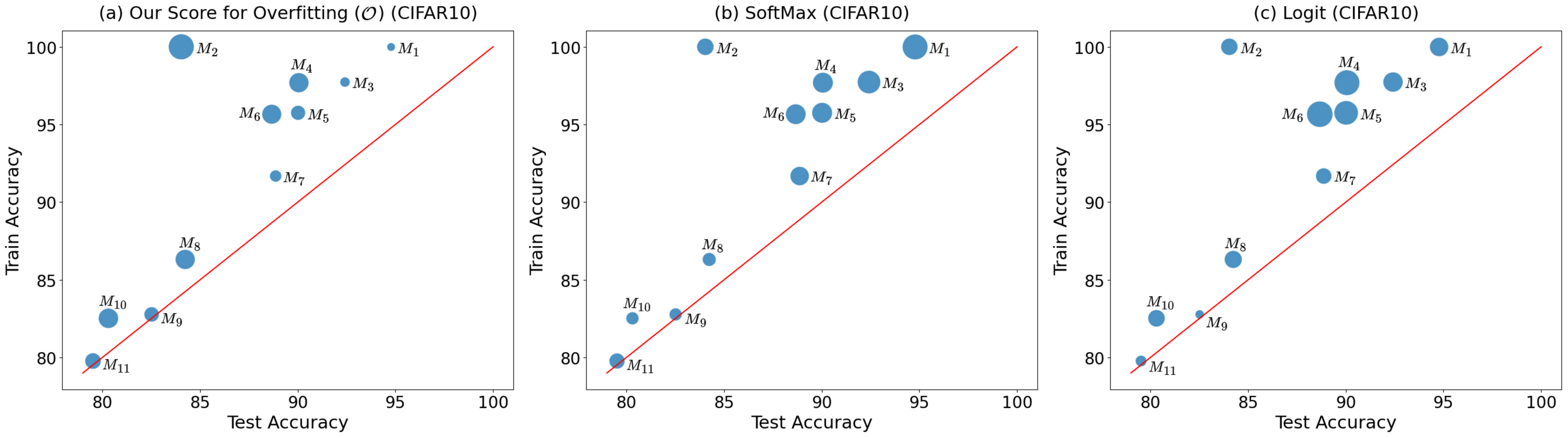}
  \caption{(a) shows the results of our method applied to the 11 different ResNet18 models that were trained on CIFAR10. The size of the circles in the plot corresponds to the degree of overfitting, as measured by the "\begin{math}\mathcal{O}\end{math}" values in Table \ref{Table:Result}. (b) and (c) show the SoftMax and Logit outputs of these models, respectively.}
  \label{fig:Network}
\end{figure}

\paragraph{CIFAR10.} The CIFAR10 dataset is a widely-used image classification dataset comprising 60,000 32x32 color images across 10 classes. To evaluate the performance of our method, we trained 11 different ResNet18 models with and without data augmentation and dropout on this dataset (with different epochs). This allowed us to obtain models with varying generalization and overfitting capabilities. For instance, in Model 8 (as shown in Fig. \ref{fig:Network} and Table \ref{Table:Result}), the first two layers of ResNet18 were utilized with data augmentation and dropout techniques (obtained in epoch 148).

\paragraph{SVHN.} The SVHN dataset is another commonly used dataset for image classification tasks consisting of 600,000 labeled digit images. For this dataset, we followed the same methodology as for CIFAR10. Results are available in Appendix \hyperref[AppendixBC]{B}.

\paragraph{CIFAR100.} The CIFAR100 dataset is similar to CIFAR10, but it differs in the number of classes and images per class. Specifically, it consists of 100 classes, with each class containing 600 images. We trained this data set on ResNet34, ResNet50, DenseNet121, VGG19, and MobileNetV2. And then evaluated our method on these models. The results are available in Appendix \hyperref[AppendixBC]{B}.

\subsection{Results} \label {Results}

\paragraph{Overfitting}

After training the 11 different ResNet18 models on CIFAR10 we randomly selected a small subset of the test data CIFAR10 and fed it into these models. To assess overfitting, we measured the values of \begin{math}\alpha\end{math} and \begin{math}\beta\end{math} for these samples and then calculated their average. The results are presented in Table \ref{Table:Result} and also plotted in Fig. \ref{fig:Network}. Our analysis shows that as the degree of overfitting increases, the sum of \begin{math}\alpha\end{math} and \begin{math}\beta\end{math} also increases, which is reflected in the "\begin{math}\mathcal{O}\end{math}" column of Table \ref{Table:Result}. As shown in Fig. \ref{fig:Network}, the size of the circles (angles) increases as we move from the bottom to the top or from right to left, indicating an increase in overfitting. Conversely, when we move diagonally from the bottom left to the top right, the size of the circles (angles) decreases (\begin{math}\mathcal{O}\end{math} decrease), indicating a decrease in overfitting.

The results for the SVHN and CIFAR100 are available in Appendix \hyperref[AppendixBC]{B}.

\begin{figure}
  \centering
  \includegraphics[width=\textwidth]{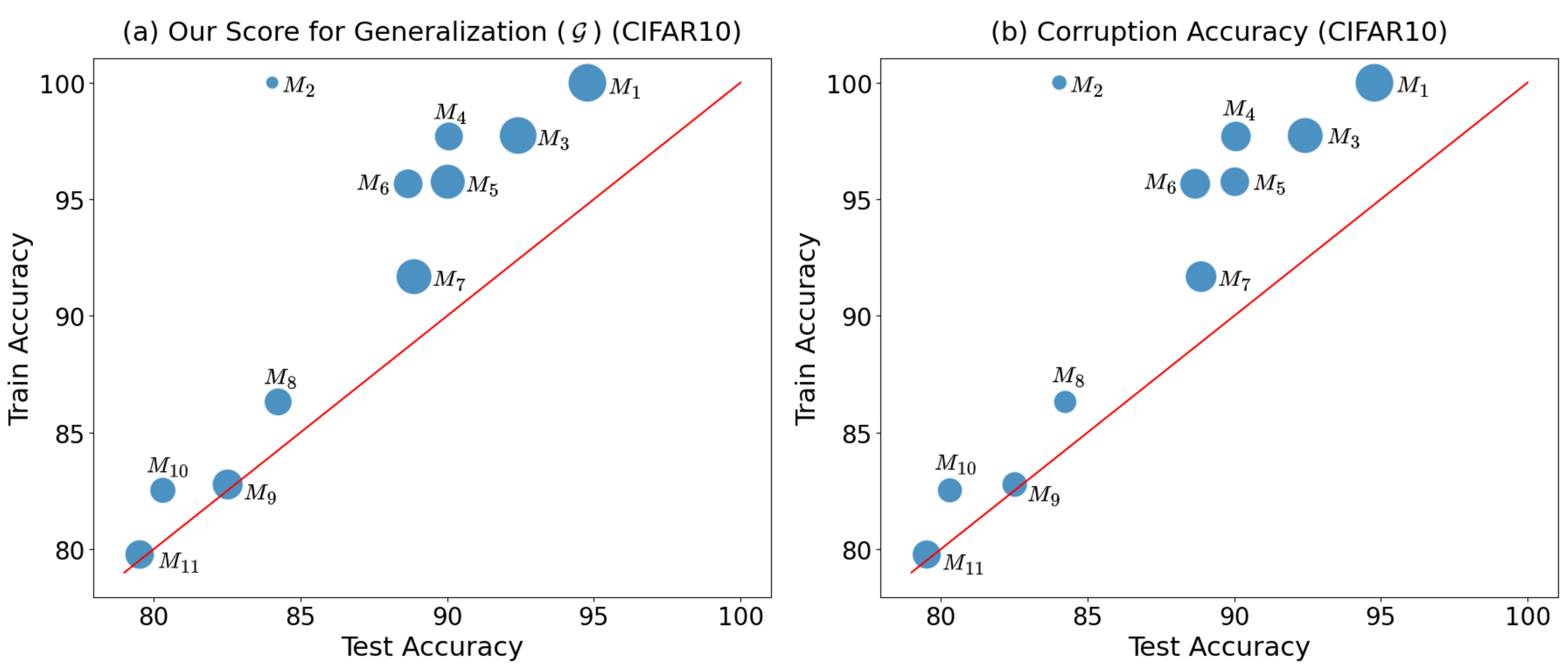}
  \caption{(a) illustrates the average accuracy across 5 different data corruptions for 11 distinct ResNet18 models trained on CIFAR10. Meanwhile, (b) showcases the outcomes of our generalization analysis method, which closely aligns with the accuracy observed for the data corruption.}
  \label{fig:Generalization}
\end{figure}

\paragraph{Generalization}

After completing the training process for 11 distinct ResNet18 models on CIFAR10, we proceeded to randomly select a small subset of the CIFAR10 test data. This subset of data was then used as input for these models to assess their generalization capability. To quantify this capability, we calculated the \begin{math}\alpha\end{math} and \begin{math}\beta\end{math} values for these samples and computed their average. Subsequently, we normalized the \begin{math}\alpha\end{math} value by dividing it by the maximum value, denoted as \begin{math}\alpha^\prime\end{math} in Table \ref{Table:Result1}. Similarly, for \begin{math}\beta\end{math}, we first calculated its absolute value and then normalized it by dividing it by the maximum value, denoted as \begin{math}\beta^\prime\end{math} in Table \ref{Table:Result1}. Our analysis revealed that as the generalization capability increased, the sum of \begin{math}\alpha^\prime\end{math} and \begin{math}\beta^\prime\end{math} also increased, as indicated in the \begin{math}\mathcal{G}\end{math} column of Table \ref{Table:Result1}. This trend is further visualized in Fig. \ref{fig:Generalization}.

To validate the effectiveness of our proposed method for analyzing generalization, we conducted a series of generalization tests. Specifically, we applied various data corruptions, including adjust\_sharpness, adjust\_brightness, gaussian\_blur, perspective, adjust\_hue, and rotate, to the test data. We then evaluated the performance of the models on these distributional shifts. The accuracy values were averaged, and are presented as the "\begin{math}Corruption\end{math}" in Table \ref{Table:Result1}. Additionally, we visualized our metric and the accuracy under these corruptions in Fig. \ref{fig:Generalization}. As you can observe, our metric aligns with the accuracy of data corruption, demonstrating the effectiveness of our method.

The results for the SVHN are available in Appendix \hyperref[AppendixBC]{B}.

\paragraph{SoftMax \& Logits.}
A common method that comes to mind for detecting overfitting is to examine the softmax or logit outputs of the model. This involves calculating the softmax or logit values for the ground truth label of each sample and then averaging these values across all samples. However, as shown in Fig. \ref{fig:Network} and Table \ref{Table:Result}, there are cases where this method fails to detect overfitting. For example, consider models 1 and 2, where the softmax or logit outputs do not indicate overfitting despite evidence of overfitting based on other metrics.

\begin{table}
  \caption{Results of training various ResNet18 on CIFAR10. The '\begin{math}\alpha\end{math}' value indicates the angle between the representation vector and the true class weight vector, while '\begin{math}\beta\end{math}' shows the angle between the representation vector and the null space. The sum of '\begin{math}\alpha\end{math}' and '\begin{math}\beta\end{math}', denoted as '\begin{math}\mathcal{O}\end{math}', indicates the degree of overfitting.}
  \label{Table:Result}
  \centering
  \begin{tabular}{lllllllllll}
    \toprule
    & \multicolumn{3}{c}{Our Method}
    & \multicolumn{2}{c}{Confidence}
    & \multicolumn{2}{c} {Accuracy}  \\
    \cmidrule(r){2-4}
    \cmidrule(r){5-6}
    \cmidrule(r){7-8}
    Model  & \hspace{2mm}\begin{math}\alpha\end{math}  & \hspace{3mm}\begin{math}\beta\end{math}  & \hspace{2mm}\begin{math}\mathcal{O}\end{math}  & \hspace{2mm}\begin{math}SoftMax\end{math}  & \hspace{1mm}\begin{math}Logits\end{math}   & \begin{math}Train\end{math}  & \begin{math}Test\end{math} & \begin{math}Difference\end{math}  \\
    \midrule
    \hspace{3mm}1   & 59.61   & -27.28   & \textbf{32.32}   & \hspace{5mm}\textbf{0.8906}  & \hspace{3mm}7.88   & 99.99   & 94.78  & \hspace{6.5mm}\textbf{5.21} \\
    \hspace{3mm}2   & 78.82   & -11.55   & \textbf{67.27}   & \hspace{5mm}\textbf{0.8078}  & \hspace{3mm}7.47   & 100.0   & 84.04  & \hspace{6.5mm}\textbf{15.96} \\
    \hspace{3mm}3   & 60.58   & -26.42   & 34.17   & \hspace{5mm}0.8649  & \hspace{3mm}8.09   & 97.73   & 92.42  & \hspace{6.5mm}5.31 \\
    \hspace{3mm}4   & 70.47   & -19.00   & 51.47   & \hspace{5mm}0.8373  & \hspace{3mm}9.53   & 97.69   & 90.06  & \hspace{6.5mm}7.63 \\
    \hspace{3mm}5   & 64.72   & -23.68   & 41.04   & \hspace{5mm}0.8377  & \hspace{3mm}9.16   & 95.75   & 90.02  & \hspace{6.5mm}5.73 \\
    \hspace{3mm}6   & 70.32   & -19.40   & 50.92   & \hspace{5mm}0.8361  & \hspace{3mm}9.71   & 95.66   & 88.67  & \hspace{6.5mm}6.99 \\
    \hspace{3mm}7   & 61.75   & -25.13   & 36.62   & \hspace{5mm}0.8240  & \hspace{3mm}7.29   & 91.68   & 88.87  & \hspace{6.5mm}2.81 \\
    \hspace{3mm}8   & 70.24   & -18.70   & 51.54   & \hspace{5mm}0.7834  & \hspace{3mm}7.61   & 86.31   & 84.24  & \hspace{6.5mm}2.07 \\
    \hspace{3mm}9   & 63.77   & -22.17   & 41.60   & \hspace{5mm}0.7782  & \hspace{3mm}6.28   & 82.77   & 82.52  & \hspace{6.5mm}0.25 \\
    \hspace{2mm}10  & 70.10   & -18.08   & 52.02   & \hspace{5mm}0.7786  & \hspace{3mm}7.50   & 82.52   & 80.31  & \hspace{6.5mm}2.21 \\
    \hspace{2mm}11  & 64.87   & -21.18   & 43.68   & \hspace{5mm}0.7978  & \hspace{3mm}6.55   & 79.77   & 79.52  & \hspace{6.5mm}0.25 \\
    \bottomrule
  \end{tabular}
\end{table}

\begin{table}
  \caption{The variable \begin{math}\alpha^\prime\end{math} represents the normalized angle between the representation vector and the target weights vector, while \begin{math}\beta^\prime\end{math} indicates the normalized absolute value of the angle between the representation vector and null space. The sum of \begin{math}\alpha^\prime\end{math} and \begin{math}\beta^\prime\end{math}, denoted by \begin{math}\mathcal{G}\end{math}, provides an indication of the model's generalization capability. The \begin{math}Corruption\end{math} represents the average accuracy across 5 different data corruptions for 11 distinct ResNet18 models trained on CIFAR10. As demonstrated, our metric aligns with the corruption accuracies observed.}
  \label{Table:Result1}
  \centering
  \begin{tabular}{lllllllll}
    \toprule
    & \multicolumn{3}{c}{Our Method}
    & \multicolumn{3}{c} {Accuracy}  \\
    \cmidrule(r){2-4}
    \cmidrule(r){5-7}
    Model  & \hspace{2mm}\begin{math}\alpha^\prime\end{math}  & \hspace{3mm}\begin{math}\beta^\prime\end{math}  & \hspace{2mm}\begin{math}\mathcal{G}\end{math}  & \begin{math}Train\end{math}  & \begin{math}Test\end{math} & \begin{math}Corruption\end{math}  \\
    \midrule
    \hspace{3mm}1   & 0.7563   & 1.0      & 1.7563  & 99.99   & 94.78  & \hspace{5mm}55.128 \\
    \hspace{3mm}2   & 1.0      & 0.4234   & 1.4234  & 100.0   & 84.04  & \hspace{5mm}35.666 \\
    \hspace{3mm}3   & 0.7686   & 0.9685   & 1.7371  & 97.73   & 92.42  & \hspace{5mm}51.896 \\
    \hspace{3mm}4   & 0.8941   & 0.6965   & 1.5906  & 97.69   & 90.06  & \hspace{5mm}46.446 \\
    \hspace{3mm}5   & 0.8211   & 0.8680   & 1.6891  & 95.75   & 90.02  & \hspace{5mm}45.584 \\
    \hspace{3mm}6   & 0.8922   & 0.7111   & \textbf{1.6033}  & 95.66   & 88.67  & \hspace{5mm}\textbf{46.716} \\
    \hspace{3mm}7   & 0.7834   & 0.9212   & \textbf{1.7046}  & 91.68   & 88.87  & \hspace{5mm}\textbf{47.51}  \\
    \hspace{3mm}8   & 0.8911   & 0.6855   & 1.5766  & 86.31   & 84.24  & \hspace{5mm}40.33  \\
    \hspace{3mm}9   & 0.8091   & 0.8127   & 1.6218  & 82.77   & 82.52  & \hspace{5mm}42.136 \\
    \hspace{2mm}10  & 0.8894   & 0.6628   & 1.5522  & 82.52   & 80.31  & \hspace{5mm}41.76  \\
    \hspace{2mm}11  & 0.8230   & 0.7764   & 1.5994  & 79.77   & 79.52  & \hspace{5mm}45.066 \\
    \bottomrule
  \end{tabular}
\end{table}

\paragraph{Angles of deeper architectures.}

During our analysis of deeper architectures on CIFAR100, we discovered that increasing the number of layers in a model (creating a deeper architecture) leads to a greater range of angles the model can achieve. For instance, let's consider a ResNet18 model and a DenseNet121 model, both exhibiting identical training and testing accuracy. In this case, the DenseNet121 model will have a higher sum of \begin{math}\alpha\end{math} and \begin{math}\beta\end{math}. We believe that this is because deeper models strive to attain superior generalization ability while mitigating the issue of overfitting.

The results supporting this observation can be found in Appendix \hyperref[AppendixBC]{B}.

\paragraph{Ablation study.} We have analyzed the size of the test set in relation to our proposed methods. This analysis reveals that the size of the test data has a minimal impact on the measures of overfitting and generalizability that we have put forward.

The evidence substantiating this observation is available in Appendix \hyperref[AppendixBC]{C}.
\section{Conclusion}
This paper addresses the issue of determining if a downloaded or received model has been overfitted without knowledge of its training accuracy or data. Overfitted models are more vulnerable to knowledge leakage, posing privacy risks. The proposed method analyzes the null space in the last layer of neural networks, quantifying overfitting and generalization using only a small subset of the testing data. The approach was evaluated on different architectures and datasets, revealing distinct patterns in the null space angle for overfitted models and poor generalization characteristics. This novel method provides insights into model vulnerability without training data, enhancing privacy and trustworthiness in deep learning models.



{

\small

\bibliographystyle{unsrt}
\bibliography{egbib}

}



\paragraph{\Large{Appendix A.}} \label{AppendixA}

\hfill
\medskip
\medskip

\textbf{Theorem 1:} Let A be an m × n matrix.

\begin{math}(i)\end{math} The null space of \begin{math}A\end{math}, \begin{math}Nul(A)\end{math}, and the row space of \begin{math}A\end{math}, \begin{math}Row(A)\end{math}, are orthogonal spaces.

\begin{math}(ii)\end{math} The left null space of \begin{math}A\end{math} and the column space of \begin{math}A\end{math}, \begin{math}{Col}(A)\end{math}, are orthogonal spaces.

\vspace{0.2mm}

\textbf{Proof:} To establish the validity of \begin{math}(i)\end{math}, let's consider a vector \begin{math}\vec{w} \in Nul(A)\end{math}, belonging to the null space of \begin{math}A\end{math}. Therefore, \begin{math}A\vec{w} = \vec{0}\end{math}. This implies that when we take the dot product of the first row of \begin{math}A\end{math} with \begin{math}\vec{w}\end{math}, we obtain 0. Similarly, by dotting the second row of \begin{math}A\end{math} with \begin{math}\vec{w}\end{math}, we get zero, and so on for each row of \begin{math}A\end{math}. Consequently, it is evident that \begin{math}\vec{w}\end{math} is orthogonal to every row of \begin{math}A\end{math}.

To demonstrate \begin{math}(ii)\end{math}, we need to consider \begin{math}\vec{w} \in Nul(A^T) \end{math} and \begin{math}\vec{b} \in Col(A) \end{math} and show that \begin{math}\vec{w}^T\vec{b} = 0\end{math}, where \begin{math}\vec{w}\end{math} and \begin{math}\vec{b}\end{math} are arbitrary vectors. Since \begin{math}A^Tw = \vec{0}\end{math}, we can apply the same reasoning as before to deduce that \begin{math}\vec{w}\end{math} is orthogonal to each row of \begin{math}A^T\end{math}. As the rows of \begin{math}A^T\end{math} correspond to the columns of \begin{math}A\end{math}, it follows that \begin{math}\vec{w}\end{math} is orthogonal to every column of \begin{math}A\end{math}, i.e., orthogonal to \begin{math}{Col}(A)\end{math}. Thus, \begin{math}(ii)\end{math} is proven.

\medskip
\medskip

\paragraph{\Large{Appendix B, C.}} \label{AppendixBC}

\begin{figure}[h]
  \centering
  \includegraphics[width = \textwidth]{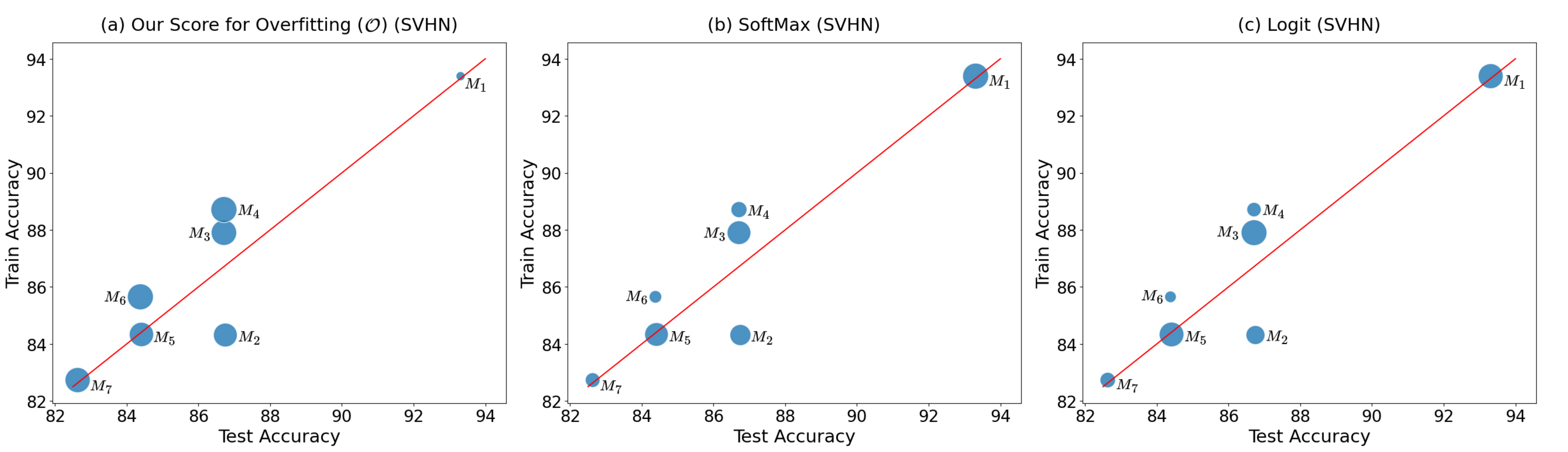}
  \caption{(a) shows the results of our method applied to the 7 different ResNet18 models that were trained on SVHN. The size of the circles in the plot corresponds to the degree of overfitting, as measured by the "\begin{math}\mathcal{O}\end{math}" values in Table \ref{Table:SVHN}. (b) and (c) show the SoftMax and Logit outputs of these models, respectively.}
  \label{fig:SVHN_overfitting}
\end{figure}

\begin{figure}[ht]
  \centering
  \includegraphics[scale=0.19]{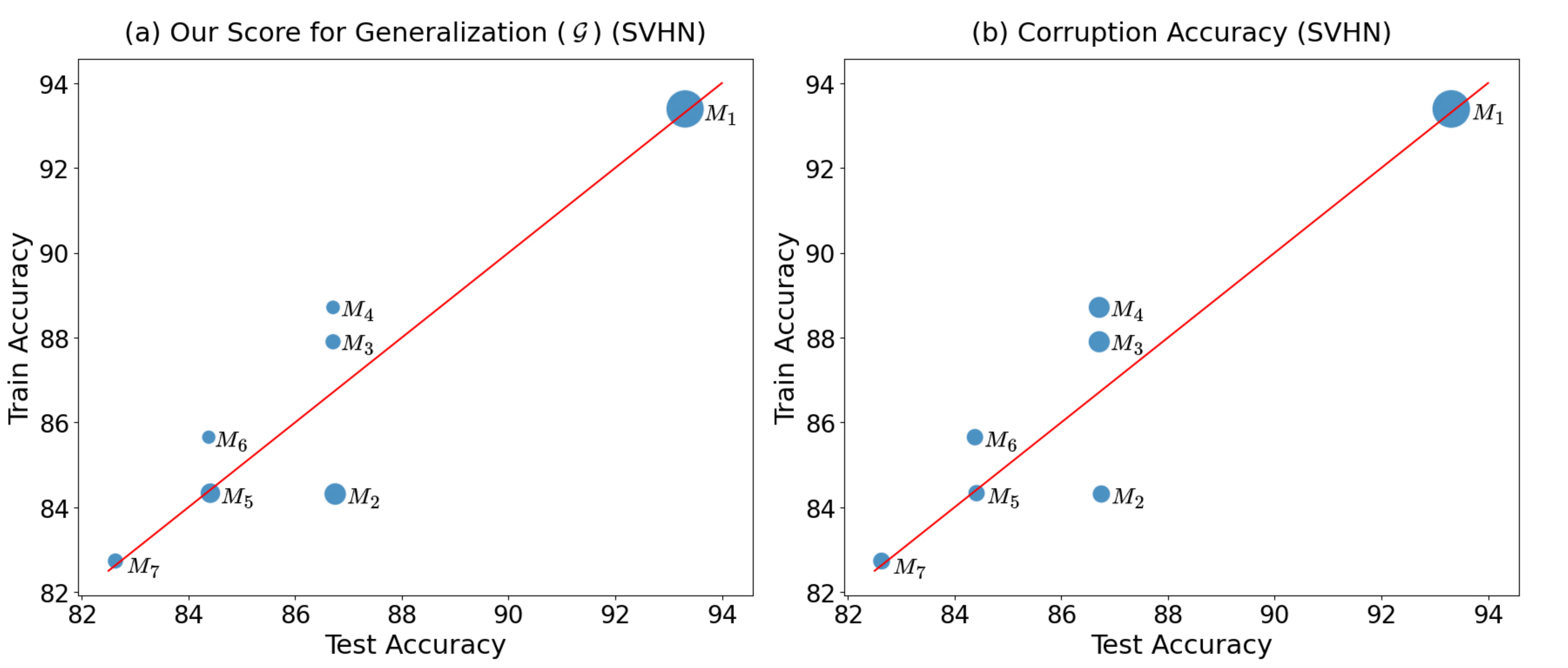}
  \caption{(a) illustrates the average accuracy across 6 different data corruptions for 7 distinct ResNet18 models trained on SVHN. Meanwhile, (b) showcases the outcomes of our generalization analysis method, which closely aligns with the accuracy observed for the data corruption.}
  \label{fig:SVHN_Generalization}
\end{figure}

\begin{figure}[ht]
  \centering
  \includegraphics[scale=0.19]{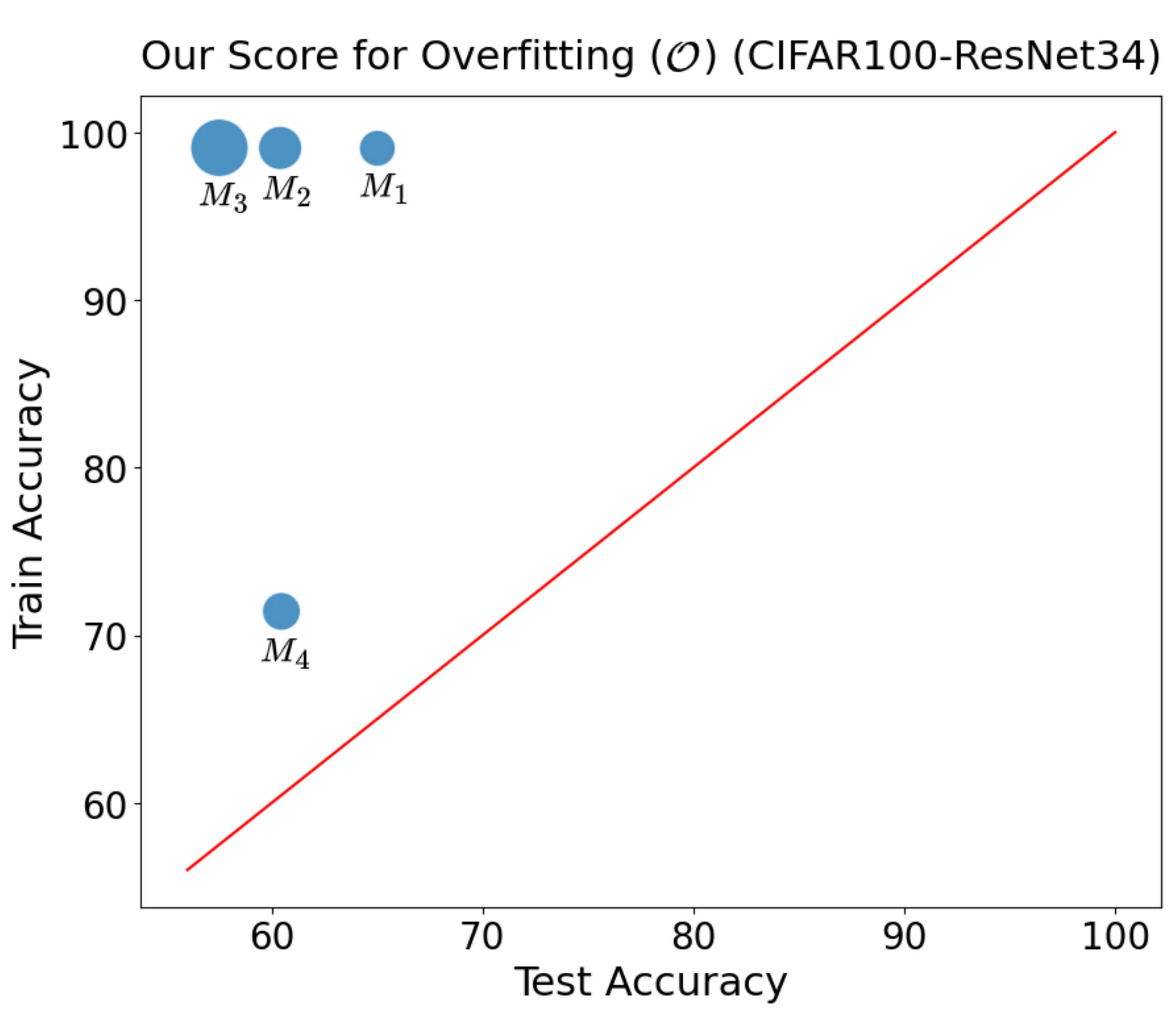}
  \caption{The figure shows the results of our method applied to the 4 different ResNet34 models that were trained on CIFAR100. The size of the circles in the plot corresponds to the degree of overfitting, as measured by the "\begin{math}\mathcal{O}\end{math}" values in Table \ref{Table:CIFAR100_ResNet34}.}
  \label{fig:CIFAR100_ResNet34}
\end{figure}

\begin{figure}[ht]
  \centering
  \includegraphics[scale=0.19]{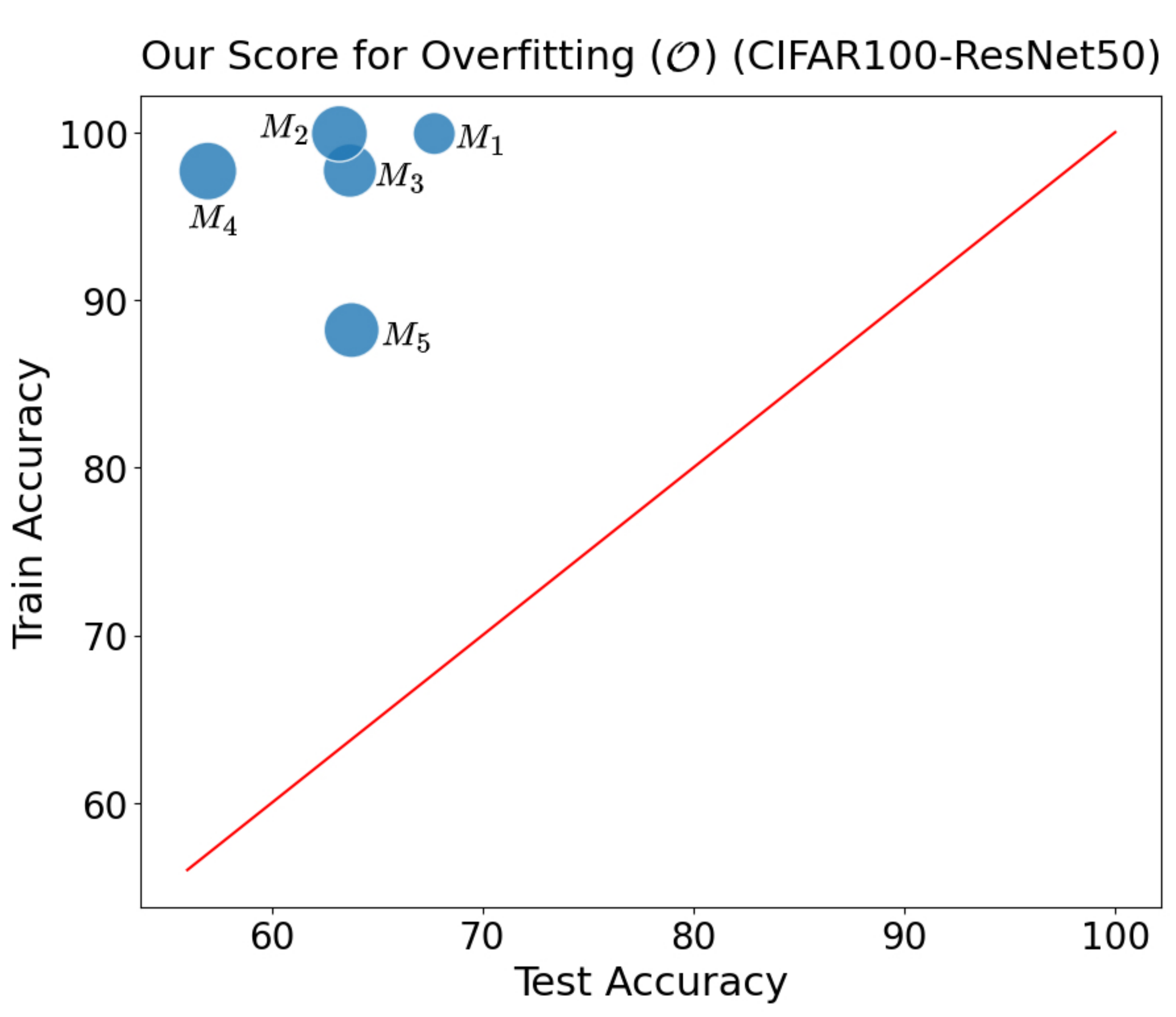}
  \caption{The figure shows the results of our method applied to the 5 different ResNet50 models that were trained on CIFAR100. The size of the circles in the plot corresponds to the degree of overfitting, as measured by the "\begin{math}\mathcal{O}\end{math}" values in Table \ref{Table:CIFAR100_ResNet50}.}
  \label{fig:CIFAR100_ResNet50}
\end{figure}

\begin{figure}[ht]
  \centering
  \includegraphics[scale=0.18]{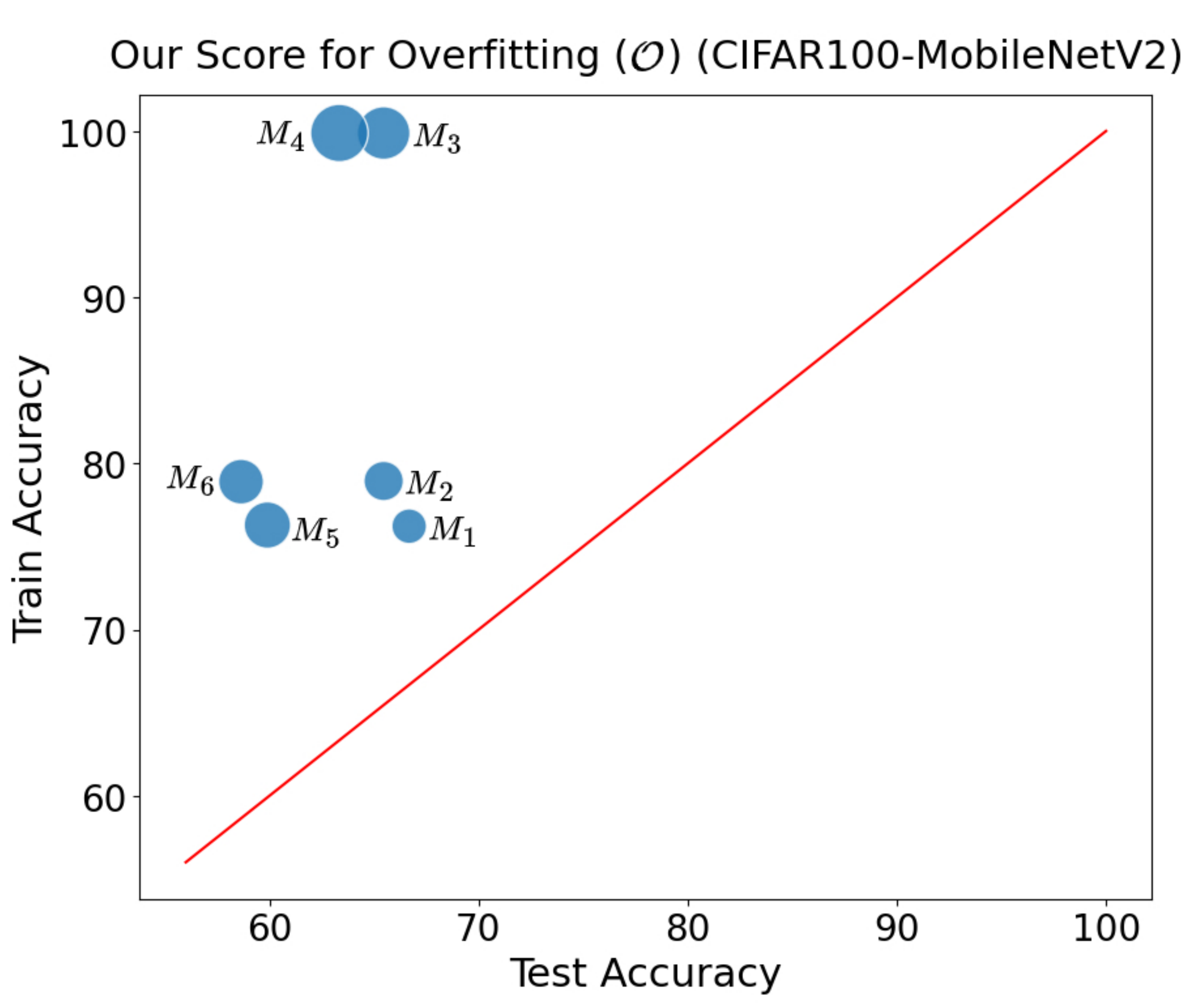}
  \caption{The figure shows the results of our method applied to the 6 different MobileNetV2 models that were trained on CIFAR100. The size of the circles in the plot corresponds to the degree of overfitting, as measured by the "\begin{math}\mathcal{O}\end{math}" values in Table \ref{Table:CIFAR100_MobileNetV2}.}
  \label{fig:CIFAR100_MobileNetV2}
\end{figure}

\begin{figure}
  \centering
  \includegraphics[scale=0.19]{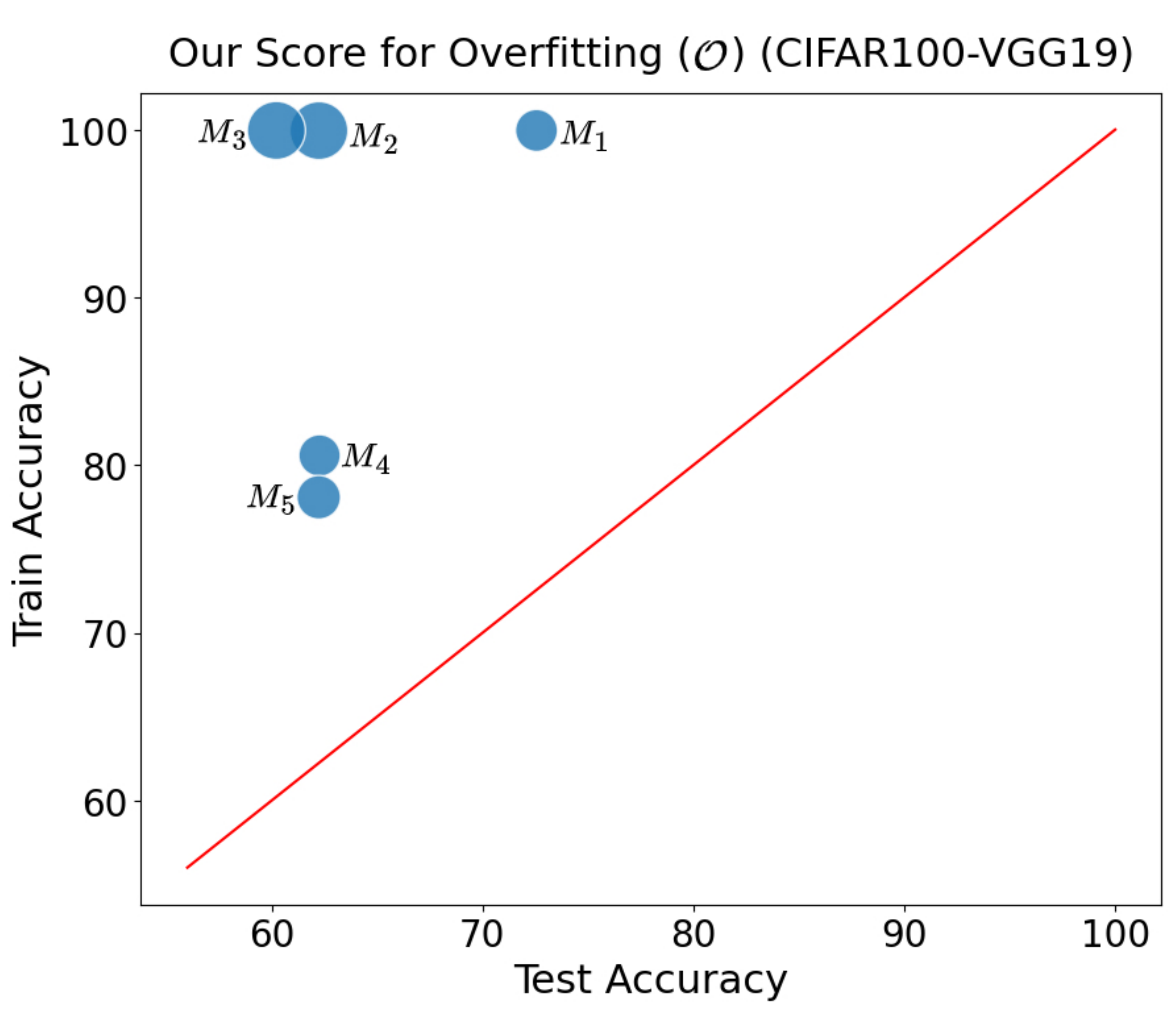}
  \caption{The figure shows the results of our method applied to the 5 different VGG19 models that were trained on CIFAR100. The size of the circles in the plot corresponds to the degree of overfitting, as measured by the "\begin{math}\mathcal{O}\end{math}" values in Table \ref{Table:CIFAR100_VGG19}.}
  \label{fig:CIFAR100_VGG19}
\end{figure}

\paragraph{Results for SVHN, and CIFAR100.} The results of training various ResNet18 models on SVHN are displayed in Figures \ref{fig:SVHN_overfitting} and \ref{fig:SVHN_Generalization}, as well as Tables \ref{Table:SVHN} and \ref{Table:SVHN_Generalization}. Additionally, the results of training various ResNet34 models on CIFAR100 can be found in Figure \ref{fig:CIFAR100_ResNet34} and Table \ref{Table:CIFAR100_ResNet34}, while the results of training various ResNet50 models on CIFAR100 are shown in Figure \ref{fig:CIFAR100_ResNet50} and Table \ref{Table:CIFAR100_ResNet50}. Furthermore, the results of training various MobileNetV2 models on CIFAR100 can be found in Figure \ref{fig:CIFAR100_MobileNetV2} and Table \ref{Table:CIFAR100_MobileNetV2}, and the results of training various VGG19 models on CIFAR100 are presented in Figure \ref{fig:CIFAR100_VGG19} and Table \ref{Table:CIFAR100_VGG19}.

\paragraph{Angles of deeper architectures.}
As you may know, the MobileNetV2 model is deeper than the ResNet50 model, with 173 layers compared to ResNet50's 161 layers. If we consider Model 2 from Table \ref{Table:CIFAR100_ResNet50} (ResNet50) and Model 4 (MobileNetV2) from Table \ref{Table:CIFAR100_MobileNetV2}, we can see that the "O" score for Model 4 is slightly higher than for Model 2.

\paragraph{Ablation study.} For the models trained on CIFAR100, only the data that were predicted incorrectly have been considered, so the size of the dataset cannot affect our method.

\begin{table}[h]
  \caption{Results of training various ResNet18 on SVHN. The '\begin{math}\alpha\end{math}' value indicates the angle between the representation vector and the true class weight vector, while '\begin{math}\beta\end{math}' shows the angle between the representation vector and the null space. The sum of '\begin{math}\alpha\end{math}' and '\begin{math}\beta\end{math}', denoted as '\begin{math}\mathcal{O}\end{math}', indicates the degree of overfitting.}
  \label{Table:SVHN}
  \centering
  \resizebox{12cm}{!}{
  \begin{tabular}{lllllllllll}
    \toprule
    & \multicolumn{3}{c}{Our Method}
    & \multicolumn{2}{c}{Confidence}
    & \multicolumn{2}{c} {Accuracy}  \\
    \cmidrule(r){2-4}
    \cmidrule(r){5-6}
    \cmidrule(r){7-8}
    Model  & \hspace{2mm}\begin{math}\alpha\end{math}  & \hspace{3mm}\begin{math}\beta\end{math}  & \hspace{2mm}\begin{math}\mathcal{O}\end{math}  & \hspace{2mm}\begin{math}SoftMax\end{math}  & \hspace{1mm}\begin{math}Logits\end{math}   & \begin{math}Train\end{math}  & \begin{math}Test\end{math} & \begin{math}Difference\end{math}  \\
    \midrule
    \hspace{3mm}1   & 63.84   & -24.36   & 39.48   & \hspace{5mm}0.7818  & \hspace{3mm}6.04   & 93.39   & 93.31  & \hspace{6.5mm}0.08  \\
    \hspace{3mm}2   & 78.08   & -12.43   & 65.65   & \hspace{5mm}0.7333  & \hspace{3mm}5.14   & 84.31   & 86.75  & \hspace{5.3mm}-2.44 \\
    \hspace{3mm}3   & 80.00   & -10.05   & 69.95   & \hspace{5mm}0.7579  & \hspace{3mm}6.19   & 87.90   & 86.71  & \hspace{6.5mm}1.19  \\
    \hspace{3mm}4   & 81.00   & -9.35    & 71.65   & \hspace{5mm}0.6940  & \hspace{3mm}4.61   & 88.71   & 86.71  & \hspace{6.5mm}2.00  \\
    \hspace{3mm}5   & 78.59   & -11.66   & 66.93   & \hspace{5mm}0.7572  & \hspace{3mm}5.99   & 84.33   & 84.41  & \hspace{5.3mm}-0.08 \\
    \hspace{3mm}6   & 80.84   & -9.33    & 71.51   & \hspace{5mm}0.6731  & \hspace{3mm}4.36   & 85.65   & 84.38  & \hspace{6.5mm}1.27  \\
    \hspace{3mm}7   & 79.43   & -10.26   & 69.17   & \hspace{5mm}0.6854  & \hspace{3mm}4.72   & 82.73   & 82.63  & \hspace{6.5mm}0.10  \\
    \bottomrule
  \end{tabular}
}
\end{table}

\begin{table}[ht]
  \caption{The variable \begin{math}\alpha^\prime\end{math} represents the normalized angle between the representation vector and the target weights vector, while \begin{math}\beta^\prime\end{math} indicates the normalized absolute value of the angle between the representation vector and null space. The sum of \begin{math}\alpha^\prime\end{math} and \begin{math}\beta^\prime\end{math}, denoted by \begin{math}\mathcal{G}\end{math}, provides an indication of the model's generalization capability. The \begin{math}Corruption\end{math} represents the average accuracy across 6 different data corruptions for 7 distinct ResNet18 models trained on SVHN. As demonstrated, our metric aligns with the corruption accuracies observed.}
  \label{Table:SVHN_Generalization}
  \centering
  \resizebox{10cm}{!}{
  \begin{tabular}{lllllllll}
    \toprule
    & \multicolumn{3}{c}{Our Method}
    & \multicolumn{3}{c} {Accuracy}  \\
    \cmidrule(r){2-4}
    \cmidrule(r){5-7}
    Model  & \hspace{2mm}\begin{math}\alpha^\prime\end{math}  & \hspace{3mm}\begin{math}\beta^\prime\end{math}  & \hspace{2mm}\begin{math}\mathcal{G}\end{math}  & \begin{math}Train\end{math}  & \begin{math}Test\end{math} & \begin{math}Corruption\end{math}  \\
    \midrule
    \hspace{3mm}1   & 0.7881   & 0.8066   & 1.5947  & 93.39  & 93.31  & \hspace{5mm}79.976 \\
    \hspace{3mm}2   & 0.9640   & 0.4116   & 1.3756  & 84.31  & 86.75  & \hspace{5mm}64.492 \\
    \hspace{3mm}3   & 0.9877   & 0.3328   & 1.3205  & 87.90  & 86.71  & \hspace{5mm}66.602 \\
    \hspace{3mm}4   & 1.00     & 0.3096   & 1.3096  & 88.71  & 86.71  & \hspace{5mm}66.486 \\
    \hspace{3mm}5   & 0.9702   & 0.3861   & 1.3563  & 84.33  & 84.41  & \hspace{5mm}64.057 \\
    \hspace{3mm}6   & 0.9980   & 0.3089   & 1.3069  & 85.65  & 84.38  & \hspace{5mm}64.085 \\
    \hspace{3mm}7   & 0.9806   & 0.3397   & 1.3203  & 82.73  & 82.63  & \hspace{5mm}64.301 \\
    \bottomrule
  \end{tabular}
  }
\end{table}

\begin{table}[ht]
  \caption{Results of training various ResNet34 on CIFAR100. The '\begin{math}\alpha\end{math}' value indicates the angle between the representation vector and the true class weight vector, while '\begin{math}\beta\end{math}' shows the angle between the representation vector and the null space. The sum of '\begin{math}\alpha\end{math}' and '\begin{math}\beta\end{math}', denoted as '\begin{math}\mathcal{O}\end{math}', indicates the degree of overfitting.}
  \label{Table:CIFAR100_ResNet34}
  \centering
  \resizebox{9cm}{!}{
  \begin{tabular}{lllllll}
    \toprule
    & \multicolumn{3}{c}{Our Method}
    & \multicolumn{2}{c} {Accuracy}  \\
    \cmidrule(r){2-4}
    \cmidrule(r){5-6}
    Model  & \hspace{2mm}\begin{math}\alpha\end{math}  & \hspace{3mm}\begin{math}\beta\end{math}  & \hspace{2mm}\begin{math}\mathcal{O}\end{math}   & \begin{math}Train\end{math}  & \begin{math}Test\end{math} & \begin{math}Difference\end{math}  \\
    \midrule
    \hspace{3mm}1   & 75.68   & -17.44   & 58.23    & 99.04    & 65.01    & \hspace{6.2mm}34.03 \\
    \hspace{3mm}2   & 75.04   & -14.95   & 60.10    & 99.06    & 60.40    & \hspace{6.2mm}38.66 \\
    \hspace{3mm}3   & 79.16   & -14.44   & 64.72    & 99.07    & 57.52    & \hspace{6.2mm}41.55 \\
    \hspace{3mm}4   & 75.70   & -17.00   & 58.70    & 71.42    & 60.46    & \hspace{6.2mm}10.96 \\
    \bottomrule
  \end{tabular}
  }
\end{table}

\begin{table}
  \caption{Results of training various ResNet50 on CIFAR100. The '\begin{math}\alpha\end{math}' value indicates the angle between the representation vector and the true class weight vector, while '\begin{math}\beta\end{math}' shows the angle between the representation vector and the null space. The sum of '\begin{math}\alpha\end{math}' and '\begin{math}\beta\end{math}', denoted as '\begin{math}\mathcal{O}\end{math}', indicates the degree of overfitting.}
  \label{Table:CIFAR100_ResNet50}
  \centering
  \resizebox{9cm}{!}{
  \begin{tabular}{lllllllll}
    \toprule
    & \multicolumn{3}{c}{Our Method}
    & \multicolumn{2}{c} {Accuracy}  \\
    \cmidrule(r){2-4}
    \cmidrule(r){5-6}
    Model  & \hspace{2mm}\begin{math}\alpha\end{math}  & \hspace{3mm}\begin{math}\beta\end{math}  & \hspace{2mm}\begin{math}\mathcal{O}\end{math}   & \begin{math}Train\end{math}  & \begin{math}Test\end{math} & \begin{math}Difference\end{math}  \\
    \midrule
    \hspace{3mm}1   & 74.44   & -17.04   & 57.39     & 99.92   & 67.71  & \hspace{6.5mm}32.21 \\
    \hspace{3mm}2   & 74.68   & -13.99   & 60.69     & 99.92   & 63.21  & \hspace{6.5mm}36.71 \\
    \hspace{3mm}3   & 75.68   & -15.63   & 60.04     & 97.71   & 63.71  & \hspace{6.5mm}34.00 \\
    \hspace{3mm}4   & 75.18   & -13.96   & 61.22     & 97.68   & 56.97  & \hspace{6.5mm}40.71 \\
    \hspace{3mm}5   & 77.24   & -16.79   & 60.45     & 88.20   & 63.79  & \hspace{6.5mm}24.41 \\
    \bottomrule
  \end{tabular}
  }
\end{table}

\begin{table}
  \caption{Results of training various MobileNetV2 on CIFAR100. The '\begin{math}\alpha\end{math}' value indicates the angle between the representation vector and the true class weight vector, while '\begin{math}\beta\end{math}' shows the angle between the representation vector and the null space. The sum of '\begin{math}\alpha\end{math}' and '\begin{math}\beta\end{math}', denoted as '\begin{math}\mathcal{O}\end{math}', indicates the degree of overfitting.}
  \label{Table:CIFAR100_MobileNetV2}
  \centering
  \resizebox{9cm}{!}{
  \begin{tabular}{lllllllll}
    \toprule
    & \multicolumn{3}{c}{Our Method}
    & \multicolumn{2}{c} {Accuracy}  \\
    \cmidrule(r){2-4}
    \cmidrule(r){5-6}
    Model  & \hspace{2mm}\begin{math}\alpha\end{math}  & \hspace{3mm}\begin{math}\beta\end{math}  & \hspace{2mm}\begin{math}\mathcal{O}\end{math}   & \begin{math}Train\end{math}  & \begin{math}Test\end{math} & \begin{math}Difference\end{math}  \\
    \midrule
    \hspace{3mm}1   & 74.73   & -20.43   & 54.30     & 76.22   & 66.68  & \hspace{6.5mm}9.54 \\
    \hspace{3mm}2   & 75.84   & -20.31   & 55.53     & 78.94   & 65.46  & \hspace{6.5mm}13.48 \\
    \hspace{3mm}3   & 75.91   & -16.03   & 59.88     & 99.88   & 65.46  & \hspace{6.5mm}34.42 \\
    \hspace{3mm}4   & 76.26   & -14.53   & 61.73     & 99.88   & 63.34  & \hspace{6.5mm}36.54 \\
    \hspace{3mm}5   & 75.03   & -17.35   & 57.68     & 76.29   & 59.90  & \hspace{6.5mm}16.39 \\
    \hspace{3mm}6   & 75.86   & -18.70   & 57.16     & 78.90   & 58.64  & \hspace{6.5mm}20.26 \\
    \bottomrule
  \end{tabular}
  }
\end{table}

\begin{table}
  \caption{Results of training various VGG19 on CIFAR100. The '\begin{math}\alpha\end{math}' value indicates the angle between the representation vector and the true class weight vector, while '\begin{math}\beta\end{math}' shows the angle between the representation vector and the null space. The sum of '\begin{math}\alpha\end{math}' and '\begin{math}\beta\end{math}', denoted as '\begin{math}\mathcal{O}\end{math}', indicates the degree of overfitting.}
  \label{Table:CIFAR100_VGG19}
  \centering
  \resizebox{9cm}{!}{
  \begin{tabular}{lllllllll}
    \toprule
    & \multicolumn{3}{c}{Our Method}
    & \multicolumn{2}{c} {Accuracy}  \\
    \cmidrule(r){2-4}
    \cmidrule(r){5-6}
    Model  & \hspace{2mm}\begin{math}\alpha\end{math}  & \hspace{3mm}\begin{math}\beta\end{math}  & \hspace{2mm}\begin{math}\mathcal{O}\end{math}   & \begin{math}Train\end{math}  & \begin{math}Test\end{math} & \begin{math}Difference\end{math}  \\
    \midrule
    \hspace{3mm}1   & 70.83   & -12.50   & 58.33     & 99.95   & 72.56  & \hspace{6.5mm}27.39 \\
    \hspace{3mm}2   & 73.56   & -11.36   & 62.20     & 99.94   & 62.24  & \hspace{6.5mm}37.70 \\
    \hspace{3mm}3   & 73.67   & -11.44   & 62.23     & 99.95   & 60.22  & \hspace{6.5mm}39.73 \\
    \hspace{3mm}4   & 70.38   & -12.16   & 58.22     & 80.57   & 62.27  & \hspace{6.5mm}18.30 \\
    \hspace{3mm}5   & 69.26   & -10.61   & 58.65     & 78.08   & 62.23  & \hspace{6.5mm}15.85 \\
    \bottomrule
  \end{tabular}
  }
\end{table}

\end{document}